# Atrial Fibrillation Detection Using Deep Features and Convolutional Networks

Sara Ross-Howe, H.R. Tizhoosh
*Kimia Lab, University of Waterloo*
Waterloo, Ontario
sara.ross/tizhoosh@uwaterloo.ca

*Abstract*—Atrial fibrillation is a cardiac arrhythmia that affects an estimated 33.5 million people globally and is the potential cause of 1 in 3 strokes in people over the age of 60. Detection and diagnosis of atrial fibrillation (AFIB) is done non-invasively in the clinical environment through the evaluation of electrocardiograms (ECGs). Early research into automated methods for the detection of AFIB in ECG signals focused on traditional biomedical signal analysis to extract important features for use in statistical classification models. Artificial intelligence models have more recently been used that employ convolutional and/or recurrent network architectures. In this work, significant time and frequency domain characteristics of the ECG signal are extracted by applying the short-time Fourier transform and then visually representing the information in a spectrogram. Two different classification approaches were investigated that utilized deep features in the spectrograms constructed from ECG segments. The first approach used a pre-trained DenseNet model to extract features that were then classified using Support Vector Machines, and the second approach used the spectrograms as direct input into a convolutional network. Both approaches were evaluated against the MIT-BIH AFIB dataset, where the convolutional network approach achieved a classification accuracy of 93.16%. While these results do not surpass established automated atrial fibrillation detection methods, they are promising and warrant further investigation given they did not require any noise pre-filtering, hand-crafted features, nor a reliance on beat detection.

## I. Introduction

Atrial fibrillation is a cardiac arrhythmia that affects an estimated 33.5 million people in the world, or approximately 0.5% of the global population [1]. This arrhythmia is characterized by rapid and irregular ventricular beats and the resultant chaotic heart rhythm is a key contributor to the occurrence of strokes. The Heart and Stroke Foundation of Canada estimates that atrial fibrillation is the cause of 1 in 3 strokes for people over the age of 60 [2], and the CDC estimates that this cardiac arrhythmia is the cause of 130,000 deaths each year in the United States [3].

Detection and diagnosis of atrial fibrillation (AFIB) is done non-invasively in clinical environments through the evaluation of electrocardiograms (ECGs). Key characteristics of AFIB in the ECG trace include a high variability in the R Peak–to-R Peak intervals (the R Peak is the apex of the electrical beat signature) and an absence of P waves (the P wave is a small wave that proceeds the QRS complex). AFIB can also be associated with an elevated ventricular rate of >100 beats per minute.

## II. Background review

Early research into automated methods for the detection of AFIB in ECG signals focused on traditional biomedical signal analysis to extract important features for use in statistical classification models. Defining work in the field was published in 2000 [4], which outlined the use of the Komolgorov-Smirnov test on cumulative distributions of delta R Peak-to-R Peak intervals (ΔRR) extracted from reference data and distributions built from 2-minute segments of ECG recordings. This method achieved a reported sensitivity of 93.2% and specificity of 96.7% for the detection of AFIB regions. In 2014, an approach was identified that used symbolic dynamics to encode ΔRR values into symbolic sequences [5]. Shannon entropy was then calculated on the symbolic sequence and a threshold was used as a cutoff to determine if the beat represented AFIB. This entropy-based method reported a sensitivity of 96.89% and specificity of 98.25% on the MIT-BIH AFIB database [5].

Research into machine learning has accelerated in recent years following the introduction of deep learning in 2015 [6]. With deep models, there are progressive layers of representation of the data, as features are automatically extracted and naturally organized into successive levels of abstraction. This allows raw data to be used directly as inputs, without significant feature engineering efforts.

Applications of deep learning towards the classification of cardiac arrhythmias in ECG signals have very recently begun to emerge. In the 2017 PhysioNet/Computing in Cardiology (CinC) Challenge, a significant number of the solutions involved deep models for the classification of cardiac rhythms. In the 2017 Computing in Cardiology conference, CinC competitors presented winning classification methods on a reference dataset that included AFIB [7,8,9,10]. The common deep network architectures shared across this work includes convolutional neural networks and recurrent neural networks [7,8,9,13,14]. Representative results from these methods demonstrated a sensitivity of 82.5% and specificity of 98.7% on the CinC dataset [10].

Hybrid combinations of recurrent and convolutional networks have also been proposed to capture the unique sequential properties and morphological features of biomedical signals [9,10,11]. Techniques are also being introduced to capture both time and frequency domain characteristics of the signals as better inputs for machine learning models. A continuous wavelet transform (CWT) was used in [12] to bound frequency information to a time scale in a window surrounding an identified beat (from the annotated reference), which was then represented by a spectrogram for subsequent processing by a convolutional neural network. This combined CWT and convolutional neural network approach achieved a sensitivity of 99.41% and specificity of 98.91% on the MIT-BIH AFIB database.

### III. DATA SET

PhysioNet is an open source repository for complex physiological signals. It contains databases for the research community to use with associated metadata and raw waveforms for ECG, gait and balance, neuroelectric, myoelectric, galvanic skin response, EMG, and respiration, amongst others [13].

The MIT-BIH AFIB database that is offered in the PhysioNet repository formed the content for the training and testing sets for this project [14]. It contains 25 patients' measurements recorded continuously over a 10-hour period. The signals were captured using ECG ambulatory recorders at Boston's Beth Israel Hospital (now the Beth Israel Deaconess Medical Center). There are two ECG lead configurations recorded for each patient (Lead I and Lead II), and the signals were captured at a 360 Hz sampling frequency and 11-bit resolution over a range of ±10 millivolts. Each ECG channel was annotated with an automated method as a first pass to identify the ECG rhythms, which include: Atrial Fibrillation, Atrial Flutter, AV Junction Rhythm, or Normal Sinus Rhythm (NSR). Corrections were then identified manually by clinical experts as well as the research community. Most of the recordings in the MIT-BIH database represent paroxysmal atrial fibrillation events. The database contained a total of 260 hours of recordings, with 94.99 hours of AFIB episodes.

#### A. Data Pre-Processing

The training and testing sets were then extracted from the Lead I ECG waveforms available in the MIT-BIH database. To determine the width of data to use as an analysis window, the accepted physiological range of heart beats per minute for an adult was used as guidance (30 bpm – 200 bpm). As AFIB is characterized by the variability of R Peak-to-Peak duration, it was hypothesized that it would be advantageous to include at least 3 beats in each analysis segment. A six second interval was thus selected, as it would provide a range of 3 beats – 20 beats in each segment. A one second stride was also used to augment the data set. During early experimentation, it was determined that it was very important to provide some balance between the AFIB samples and the NSR samples (plus other cardiac arrhythmias) in each training batch. As the episodes of AFIB are predominately paroxysmal, if the data segments were sequentially organized the networks would be shown many "normal" examples before encountering ones with AFIB. Samples were assigned a binary classification as "AFIB" or "Not-AFIB" if any beats in the analysis region were annotated with the arrhythmia. The data was randomly assigned to a training and testing set based on a 70% training and 30% testing allotment. After data augmentation, a total of 407,577 training samples and a corresponding 174,573 testing samples were generated. This represents the equivalent of 679.30 recording hours for training and 290.96 recording hours for testing.

### IV. PROPOSED SOLUTION

The proposed solution sought to represent important time and frequency domain characteristics of the ECG signal by applying the short-time Fourier transform and rendering the result in a spectrogram. The spectrogram could then leverage established deep learning architectures developed through research in computer vision. Two classification methods were investigated with the first using the pre-trained DenseNet model to extract features from the spectrograms and Support Vector Machines for classification. The second approach used the spectrogram as a direct input into a convolutional network. The ultimate goal of this research is to establish a framework for analyzing any biomedical signal through techniques that did not explicitly rely on signal pre-processing nor hand-crafted feature engineering.

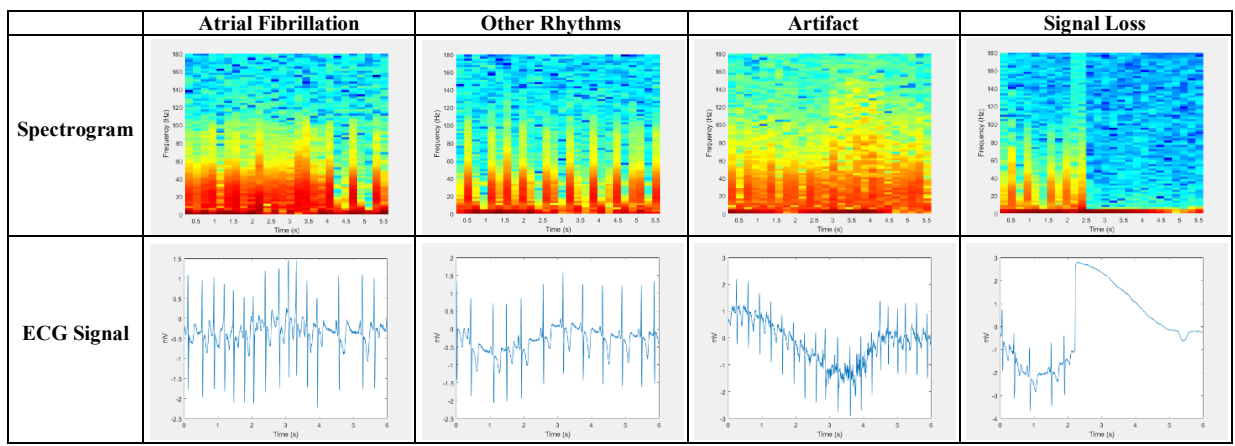

FIGURE 1. SPECTROGRAMS FOR ECG SIGNALS

*A. Spectrograms*

The ECG is a non-stationary signal that contains frequency content that changes over time. Important information regarding the ECG signal can thus be represented in both the time and frequency domains. To simultaneously represent both feature domains, spectrograms were generated for each ECG segment using the short-time Fourier transform (STFT). In equation 1, the STFT is represented for signal x[n] over window w[n]. A segment of 150 samples was used to calculate the spectrum, with a 50% overlap between successive windows. This segmentation strategy provided the best visual beat isolation during experimentation. The sequential spectrum information is then visually encoded as a progressive column in a resultant image. The magnitude of the spectrum at each frequency is represented by a colormap. In Figure 1, ECG trace segments with different rhythm characteristics are illustrated along with the resultant spectrograms. The spectrogram is represented in equation 2 below as the magnitude squared of the STFT.

$$X(\tau, \omega) = \sum_{n=-\infty}^{\infty} x[n] w[n-m] e^{-j\omega t} \quad (1)$$

$$|X(\tau, \omega)|^2 \quad (2)$$

*B. Feature Extraction*

The densely connected convolutional network (DenseNet) was introduced at CVPR in 2017 [15] and this architecture leveraged the observations that convolution networks could be deeper and more efficient to train if they contained shorter connection paths between the input and output layers. The DenseNet structure provided direct connections between each successive layer in the model and it produced state-of-the art object recognition performance against benchmark data sets such as CIFAR-10, CIFAR-100, SVHN, and ImageNet. The ImageNet database currently contains over 14 million images that are organized according to the WordNet object classification hierarchy [16]. DenseNet models are offered through *Keras* that have been trained for classification on the ImageNet database [17]. The 201-layer version of DenseNet was selected for this research and the features used for classification are extracted from the output of the final average pooling layer (1x1920).

*C. Classifier*

The DenseNet features are extracted from the spectrograms and then used as inputs into a final classifier to determine whether the ECG segment contains AFIB or other (including NSR). Support Vector Machines (SVM) were used as the classifier and were configured with a linear boundary (LinearSVC) [18]. A non-linear kernel (radial basis function) was investigated for this problem as well, however, due to the larger size of the data set it was difficult to train. A convolutional network architecture was also investigated that used 128x64 grayscale spectrograms as inputs (see Figure 2). This network contained two convolutional layers with 5x5 convolutional filters and subsequent 2x2 average pooling filters. Two fully connected network layers processed the outputs of the convolutional layers and fed into two classification nodes. Batch normalization was applied to the outputs of each layer, which helped to increase network training and produced an overall lower classification error [19]. Weights in all layers were initialized to small values in proportion to the number of nodes in the layer ($min = \frac{-1}{\sqrt{n_{inputs}}}$, $max = \frac{1}{\sqrt{n_{inputs}}}$). The loss function used was the mean squared error and soft max with logits was used on the outputs. The ReLU activation function was used throughout the network and the Adam Optimizer allowed the network to converge faster with lower loss [20]. Dropout was also used in the fully connected layers (3 and 4) to support a better generalization.

V. EXPERIMENTS AND RESULTS

The two classification approaches were evaluated against a 70%-30% training/testing split on the extracted MIT-BIH data set. The Spectrogram+DenseNet+SVM approach was evaluated against a random 5-fold cross validation and the Spectrogram+ConvNet approach was evaluated against a single random split and trained over 100 epochs. Computations were run on an Amazon Web Service virtual instance, configured as a Windows Server 2016 machine with 4 CPUs and 64 GB of default memory, and a single Tesla K80 GPU. Tensorflow 1.4.1 was used with GPU processing enabled.

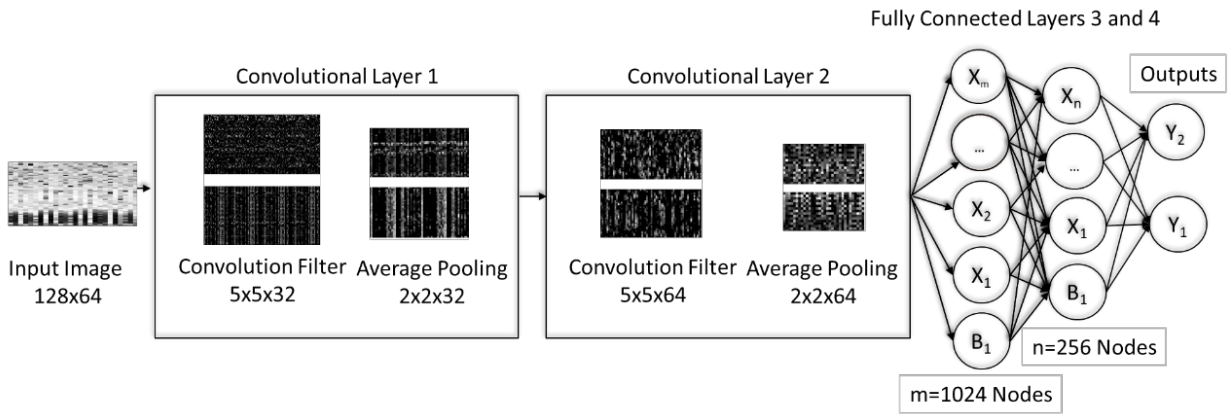

FIGURE 2. CONVOLUTIONAL NETWORK ARCHITECTURE FOR ATRIAL FIBRILLATION DETECTION

| Performance of Classification Methods for Atrial Fibrillation Detection | | | | | |
|---|---|---|---|---|---|
| *Authors* | *Year* | *Data Set* | *Sensitivity (%)* | *Specificity (%)* | *Accuracy (%)* |
| Tatano and Glass | 2000 | MIT-BIH AFB | 93.2 | 96.7 | 94.95 |
| Zhou et al | 2014 | MIT-BIH AFB | 96.89 | 98.25 | 97.57 |
| Limam et al | 2017 | CinC | 82.5 | 98.7 | 90.6 |
| Runnan et al | 2018 | MIT-BIH AFB | 99.41 | 98.9 | 99.16 |
| Spectrogram + DenseNet + SVM | 2018 | MIT-BIH AFB | 88.38±0.02 | 95.14±0.01 | 92.18±0.48 |
| Spectrogram + ConvNet | 2018 | MIT-BIH-AFB | 98.33 | 89.74 | 93.16 |

TABLE 1. PERFORMANCE OF CLASSIFICATION METHODS FOR ATRIAL FIBRILLATION DETECTION

Results for the experiments are presented in Table 1 along with a comparison to other published AFIB detection approaches. The combination Spectogram+DenseNet+SVM achieved a sensitivity of 88.38±0.02%, specificity of 95.14±0.01%, and accuracy of 92.18±0.48%. The Spectrogram+ConvNet method achieved a better accuracy at 93.16% and sensitivity of 98.33%, but with a lower specificity of 89.74%.

VI. CONCLUSIONS AND FUTURE WORK

This research proposed a method to combine time and frequency characteristics of ECG signals by applying the short-time Fourier transform and visually representing the data as a spectrogram. Two different classification approaches were investigated that utilized deep features from the spectrograms constructed from six second ECG segments. The first approach used a pre-trained DenseNet model to extract features that were then classified using SVM, and the second approach used the spectrograms as direct input into a convolutional network. The convolutional network demonstrated better classification performance than the DenseNet+SVM approach with an overall accuracy of 93.16%. Neither approach surpassed the established atrial fibrillation detection methods but were promising techniques that warrant further study as they did not require any traditional noise pre-filtering, hand-crafted features, nor relied on beat detection mechanisms. To improve the diagnostic accuracy of the ConvNet models, stacked denoising autoencoders will be further investigated as integrated pre-processing layers to remove confounding noise such as movement artefacts. Transfer learning is also an area that will be further studied with both the DenseNet and convolutional network models potentially benefitting from additional pre-training against other ECG data sets.